\documentclass[sigconf]{acmart}

\DeclareUnicodeCharacter{FF0C}{,}

\AtBeginDocument{%
  \providecommand\BibTeX{{%
    \normalfont B\kern-0.5em{\scshape i\kern-0.25em b}\kern-0.8em\TeX}}}

\usepackage{booktabs}
\usepackage{multirow}
\usepackage{threeparttable}
\usepackage{comment}
\usepackage{subcaption}
\usepackage{xspace}
\usepackage{pifont}
\usepackage{amsmath}

\begin{document}

\title{3DGesPolicy: Phoneme-Aware Holistic Co-Speech Gesture Generation Based on Action Control}



\author{Xuanmeng Sha}
\affiliation{%
  \institution{The University of Osaka}
  \country{Japan}
}
\email{shaxuanmeng@gmail.com}

\author{Liyun Zhang}
\affiliation{%
  \institution{The University of Osaka}
  \country{Japan}
}
\email{liyun.zhang@lab.ime.cmc.osaka-u.ac.jp}

\author{Tomohiro Mashita}
\affiliation{%
  \institution{Osaka Electro-Communication University}
  \country{Japan}
}
\email{mashita@osakac.ac.jp}

\author{Naoya Chiba}
\affiliation{%
  \institution{The University of Osaka}
  \country{Japan}
}
\email{chiba@nchiba.net}

\author{Yuki Uranishi}
\affiliation{%
  \institution{Osaka University}
  \country{Japan}
}
\email{yuki.uranishi.cmc@osaka-u.ac.jp}



\begin{abstract}
  Generating holistic co-speech gestures that integrate full-body motion with facial expressions suffers from semantically incoherent coordination on body motion and spatially unstable meaningless movements due to existing part-decomposed or frame-level regression methods, We introduce 3DGesPolicy, a novel action-based framework that reformulates holistic gesture generation as a continuous trajectory control problem through diffusion policy from robotics. By modeling frame-to-frame variations as unified holistic actions, our method effectively learns inter-frame holistic gesture motion patterns and ensures both spatially and semantically coherent movement trajectories that adhere to realistic motion manifolds. To further bridge the gap in expressive alignment, we propose a Gesture-Audio-Phoneme (GAP) fusion module that can deeply integrate and refine multi-modal signals, ensuring structured and fine-grained alignment between speech semantics, body motion, and facial expressions. Extensive quantitative and qualitative experiments on the BEAT2 dataset demonstrate the effectiveness of our 3DGesPolicy across other state-of-the-art methods in generating natural, expressive, and highly speech-aligned holistic gestures.
\end{abstract}

\ccsdesc[500]{Computing methodologies~Computer vision}
\ccsdesc[500]{Computing methodologies~Vision for robotics}
\ccsdesc[500]{Computing methodologies~Neural networks}






\keywords{Co-speech generation, Gesture generation, Diffusion Policy, Trajectory control, 3D Human Animation}





\maketitle

\section{Introduction}
\label{sec:intro}

\begin{figure}[t]
  \centering
  \includegraphics[width=\linewidth]{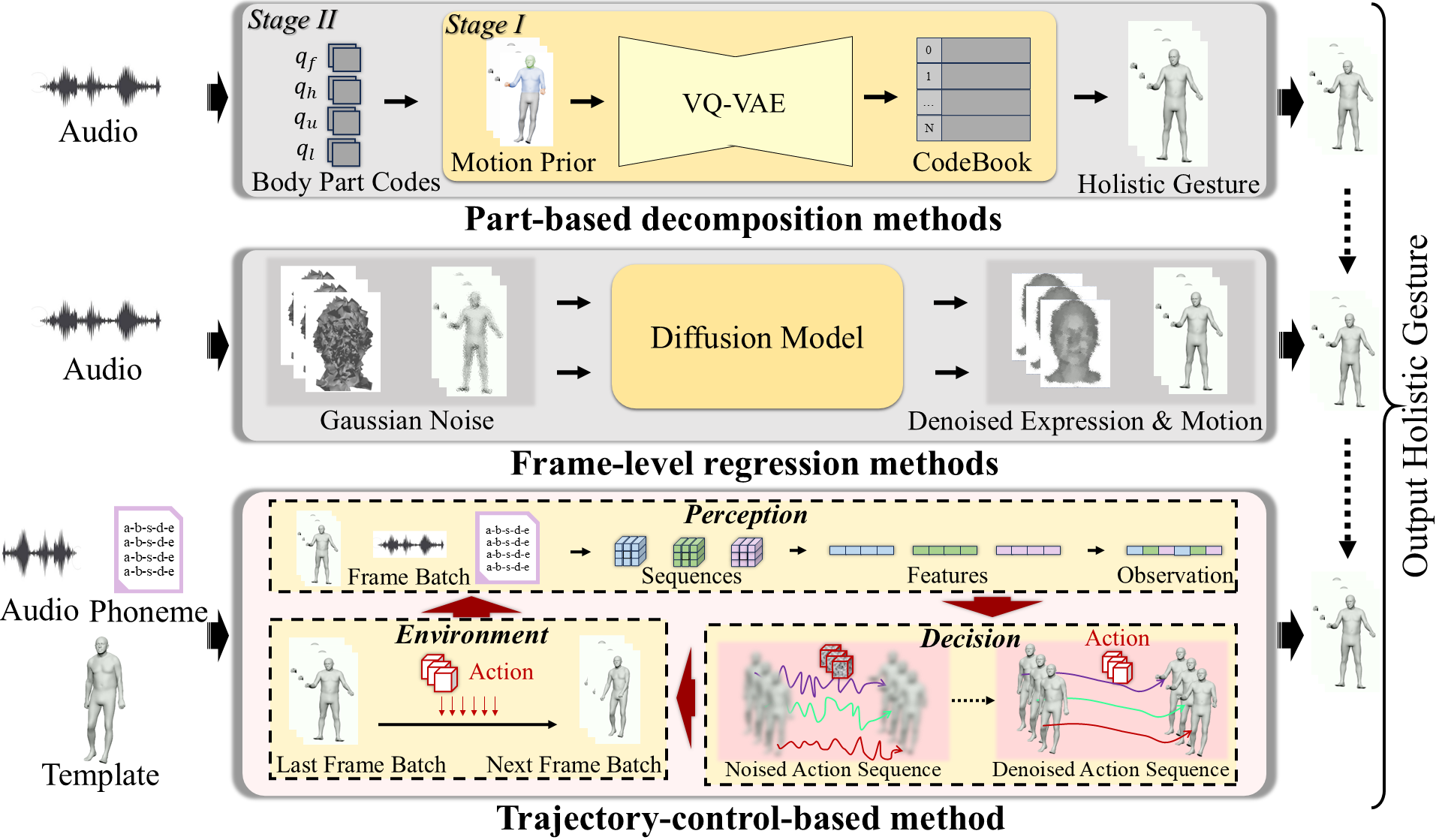}
  \caption{Comparison of holistic co-speech gesture generation paradigms. Unlike part-based or frame-level methods that use discrete codebooks or absolute gesture regression, 3DGesPolicy reformulates the task as an action-based trajectory control problem. By refining unified gesture actions in phoneme level through a diffusion policy training loop, our framework ensures stable, spatially and semantically coherent movements.}
  \label{fig1}
\end{figure}

Realistic 3D digital humans are essential for applications such as virtual assistants and social robots to AR/VR telepresence, where motion quality directly affects user engagement and communication effectiveness \cite{alexanderson2023listen,zhi2023livelyspeaker,xu2024mambatalk}. Central to this goal is co-speech gesture synthesis, while speech-driven isolated body gestures \cite{liu2022disco,liu2022beat,bhattacharya2021text2gestures,qi2024emotiongesture} and talking faces \cite{fan2022faceformer,xing2023codetalker,sha20253dfacepolicyaudiodriven3dfacial} have achieved impressive results, holistic co-speech gesture synthesis integrating both remains difficult due to challenges in body-part coordination and semantic alignment \cite{liu2024emage,xu2024mambatalk}.

Existing methods primarily follow two approaches. As illustrated in Figure~\ref{fig1}, part-based decomposition methods \cite{liu2024emage,xu2024mambatalk} utilize discrete codebooks \cite{geng2023human} to predict body-part motion. While this design provides strong motion priors, quantized body segments weaken inter-part coordination and motion details, resulting in semantically incoherent gestures \cite{kingma2013auto}. In contrast, frame-level regression methods \cite{chen2024diffsheg,zhou2025exges} directly regress gesture sequences with diffusion models, preserving motion detail and diversity. However, they are sensitive to noise and heavily dependent on training data distributions \cite{zhou2025exges}, lacking robustness in capturing speech-aligned motion and tending to generate semantically meaningless movements. Moreover, both paradigms often neglect spatial movement and interaction in 3D space, leading to unnatural and unrealistic results with unstable and floating motions.

To address these limitations, we introduce 3DGesPolicy, a novel trajectory-control framework for holistic co-speech expressive gesture generation using diffusion policy mechanisms \cite{chi2025diffusion} from robotic control. Unlike prior work that predicts absolute gestures or discrete body codes, we model frame-to-frame variations of full-body and facial motion as unified holistic actions and learn a visuomotor policy that generates gesture movement trajectories conditioned on current motion state, audio, and phoneme-level semantics. This reformulation shifts the task from gesture regression to trajectory control, enabling the model to learn semantic-aware movement patterns in 3D space rather than merely fitting to absolute gesture poses or decomposed body parts. This trajectory-control approach ensures stable and grounded motion that naturally follows physical constraints, producing more realistic results while maintaining robustness across diverse speech inputs.

To ensure semantic alignment and eliminate meaningless movements, we design a novel Gesture-Audio-Phoneme (GAP) fusion module that deeply integrates multi-modal signals. The GAP module employs separate encoders for audio and phoneme features, then aligns and fuses them with gesture representations through a cross-modal attention mechanism. This design ensures that generated actions are directly guided by linguistic semantics at the phoneme level, preventing the production of excessive or irrelevant movements while promoting speech-aligned coordination between body motion and facial expressions, achieving fine-grained phoneme-level synchronization that produces natural, semantically meaningful, and emotionally expressive results.

Extensive experiments demonstrate that the effectiveness of 3DGesPolicy compared with other state-of-the-art methods in generating natural and expressive gestures across both body and facial regions. Our main contributions are three-fold:
\begin{itemize}
    \item \textbf{We propose 3DGesPolicy, a trajectory-control framework that reformulates holistic gesture generation as an action-based diffusion policy problem.} By modeling frame-to-frame variations as unified actions, our method learns stable and natural movement trajectories that naturally follow speech dynamics.
    \item \textbf{We propose a Gesture-Audio-Phoneme (GAP) fusion module that establishes fine-grained phoneme-level alignment across modalities.} Through adaptive gating and cross-modal attention, this module deeply integrates speech acoustics, linguistic semantics, body motion, and facial expressions.
    \item \textbf{Extensive experiments validate the effectiveness and superiority of our approach for holistic gesture generation. } Both quantitative and qualitative evaluations demonstrate that our method achieves natural and stable full-body motion with precise lip-speech synchronization and semantically meaningful gestures compared with other state-of-the-art methods.
\end{itemize}


\section{Related Work}
\label{sec:RelatedWork}

\subsection{Holistic Co-Speech Gesture Generation}
Co-speech gesture generation aims to create full-body motions that align with speech and emotion. While recent studies focus on analyzing subtle emotional dynamics in dialogue \cite{MicroEmo-arxiv, MicroEmo-mm}, generative approaches aim to synthesize these cues. Early research included rule-based methods \cite{kopp2004synthesizing,wagner2014gesture} and data-driven approaches \cite{liu2022disco,liu2022beat,yoon2019robots,bhattacharya2021text2gestures,qi2024emotiongesture}. Recently, holistic generation has emerged to coordinate body pose, facial expressions, and lip movements \cite{yi2023generating,ng2024audio,liu2024emage,xu2024mambatalk}. Current methods mainly follow two paradigms: part-based decomposition and frame-level regression.

\subsubsection{Part-Based Decomposition Methods.}
These methods use discrete representations of body parts to generate gestures. TalkSHOW \cite{yi2023generating} used compositional VQ-VAEs \cite{van2017neural} to encode body and hand motions. EMAGE \cite{liu2024emage} introduced masked audio-gesture modeling for better reconstruction. MambaTalk \cite{xu2024mambatalk} integrated selective state space models to improve efficiency. 
However, discrete quantization often discards fine-grained details \cite{kingma2013auto}. Furthermore, separating body parts can weaken coordination, leading to unnatural holistic motion.

\subsubsection{Frame-Level Regression Methods.}
These methods use diffusion models to regress gestures in continuous space, offering higher diversity. DiffSHEG \cite{chen2024diffsheg} modeled the joint distribution between faces and bodies. ExGes \cite{zhou2025exges} used retrieval-enhanced frameworks and contrastive learning \cite{chen2020simple,Sha:20243p} to identify keyframes. Nevertheless, since these methods predict absolute poses, they are sensitive to noise and lack robustness. This often results in meaningless movements that deviate from realistic motion manifolds.

\subsection{Diffusion Policy Model}
Diffusion policy \cite{chi2025diffusion} is a powerful robotics framework for learning complex manipulation tasks from visual observations. Advanced visual perception and translation techniques \cite{Panoptic-wacv, Panoptic-tcsvt, Thermal-to-Color, uneven} can further enhance the robustness of such observational encodings. It has been applied to reinforcement learning \cite{levine2016end,lee2020learning}, imitation learning \cite{rajeswaran2017learning,zeng2021transporter}, and motion planning \cite{ichter2018learning,florence2022implicit,10.1007/978-3-030-27526-6_50}. Recent works have added multi-view conditioning and optimization to this framework \cite{ke20243d,ren2024diffusion}. Notably, 3D Diffusion Policy \cite{ze20243d} decoupled perception and decision-making for better 3D task performance. Inspired by these robot control successes, we apply the diffusion policy paradigm to gesture generation to achieve stable and natural trajectory control.


\section{Methodology}
\label{sec:Methods}

\begin{figure*}[t]
  \centering
  \includegraphics[width=0.9\linewidth]{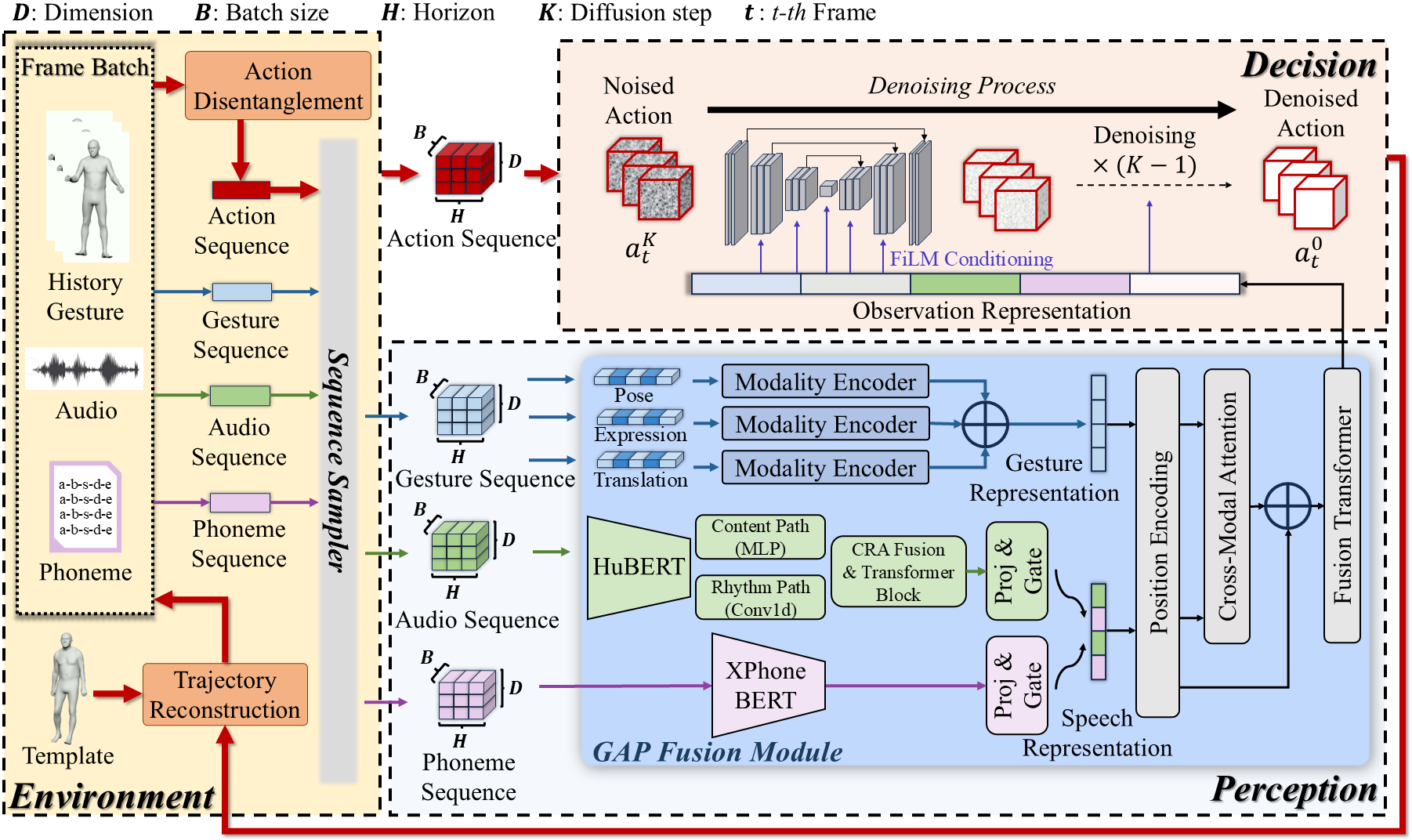}
  \caption{\textbf{Overview of 3DGesPolicy architecture. Our architecture first disentangles history gesture into action sequences, then Perception module encodes gesture, audio and phoneme sequences into observation representations that serve as conditions for Decision module with GAP fusion module, where actions are produced through denoising and drive the template to output following holistic gestures.}}
  \label{method}
\end{figure*}

\subsection{Problem Formulation}
We reformulate holistic co-speech gesture generation as a continuous trajectory control problem. In this paradigm, we aim to learn a gesture action policy (\text{3DGesPolicy}) that drives a 3D human model template to perform smooth, natural, expressive, and speech-aligned body movements and facial expressions.

Given the holistic gesture $g^{1:N} = (g^1, g^2, ..., g^N) \in \mathbb{R}^{55\times6+100+4+3}$ including 55 pose joints in Rot6D, $\mathbb{R}^{100}$ FLAME parameters, $\mathbb{R}^{4}$ foot contact labels, $\mathbb{R}^{3}$ global translations for $N$ frames, we define the temporal differential gesture deviation as the action $a^{1:N}_0 = (a^1_0, a^2_0, ..., a^N_0) \in \mathbb{R}^{N \times D_a}$, where $D_a$ is the action dimension. This action space naturally ensures motion smoothness and mitigates the jittering issues common in absolute position prediction. Conditioning on audio context $s^{1:N}$, phoneme signal $m^{1:N}$ and historical gesture states $g_h$, the Gaussian noise action sequence $a^{1:N}_k$ is iteratively refined into a clean action sequence $a^{1:N}_0$, where $k \in \{1,...,K\}$ is the diffusion step. Therefore, the goal of our framework is to learn a diffusion policy that predicts the optimal action sequence $a_0$ conditioned on the multimodal observation $O \in \{O_s, O_m, O_g\}$ with audio, phoneme, and current gesture. The problem can be formulated as:
\begin{equation}
    a^{1:N}_0 = \text{3DGesPolicy}(a^{1:N}_K, O, K),
\end{equation}
where $K$ denotes the total diffusion steps. With the predicted action sequence $a^{1:N}_0$ and the model gesture template $g_{temp}$, the holistic gesture $g^n_0$ at the $n$-th frame is presented as:
\begin{equation}
    g^n_0 = g_{temp} + \sum_{i=1}^{n} a^i_0 \cdot \Delta t, \quad n \in \{1, ..., N\},
\end{equation}
where $\Delta t$ represents the time step duration. By predicting the action, our method generates continuous, smooth, and dynamic trajectories of holistic gestures.


\subsection{Architecture}

\subsubsection{Overview.}
We design our framework as a closed-loop policy learning system in robotics and comprise three modules: Environment, Perception, and Decision. As illustrated in Figure~\ref{method}. The Environment module constructs the action space by disentangling raw gesture sequences and samples multimodal inputs of gesture, audio and phoneme into fixed-length temporal windows, establishing a stable foundation for policy learning. The Perception module encodes historical gesture sequence and synchronized audio and phoneme sequences into a comprehensive observation representation through our proposed GAP fusion module. The Decision module stands an important role for the gesture's action policy training through a conditional diffusion process, it iteratively refines noisy action sequences into clean gesture trajectories guided by observation representation from Perception module. Finally, the action is accumulated on the 3D human model template to reconstruct the holistic gestures. This architecture design enables our model to learn smooth and expressive gesture control policies that naturally follow speech rhythms and semantics.

\subsubsection{Environment.}
The Environment module constructs the action space and prepares multimodal data for efficient policy training. By disentangling raw gesture sequences into learnable temporal variations and organizing variable-length inputs into fixed-horizon batches, it ensures the policy is trained within a concise, information-rich context, promoting motion consistency and training stability.

\textit{Action Formulation:}
A critical component of our approach is the formulation of action sequences that effectively capture holistic gesture dynamics spanning both body motion and facial expressions. We define actions as temporal differential representations that encode kinematic variations between consecutive frames with adaptive scaling mechanisms. This motion-centric formulation enables our model to learn smooth gesture movement trajectories while naturally preserving temporal continuity.

Given a holistic gesture sequence $g^{1:N} = \{g^1, g^2, ..., g^N\} \in \mathbb{R}^{N \times D_g}$, we first compute the fundamental temporal displacement operator:
\begin{equation}
    \mathcal{D}_{temporal}^{n}  = g^{n+1} - g^{n}, n\in[0,N-1]
\end{equation}
This operator captures the raw inter-frame motion vectors that serve as the basis for action construction. To enhance motion sensitivity and training stability, we apply an adaptive scaling mechanism that modulates the action magnitude based on motion intensity. The scaling factor is defined as:
\begin{equation}
    \Lambda_{adaptive}  = \exp(-\beta\cdot\| \mathcal{D} _{temporal}^{n} \| _{F}^{2}) 
\end{equation}
The exponential function provides natural motion-based weighting where $\beta$ controlling the sensitivity decay rate. This ensures that subtle movements receive higher scaling weights while preventing large motions from dominating the learning process. $\| \cdot \| _{F}$ denotes the Frobenius norm. The final action formulation is expressed as:
\begin{equation}
    a^{n}  = \varepsilon_{scaling}\cdot \Lambda_{adaptive} \cdot \mathcal{D}_{temporal}^{n}
\end{equation}
where $\varepsilon_{scaling}$ is the base scaling parameter. This differential representation naturally preserves motion continuity through temporal accumulation during trajectory reconstruction.

After computing actions, the synchronized sequences are partitioned into fixed-duration segments called horizons $H$ to provide a locally consistent temporal context with sequence sampler. These small segments prepare the multi-modal data for the Perception module to encode historical context and for the Decision module to refine gesture trajectories.

\subsubsection{Perception.}
The Perception module transforms multimodal observations within the observation horizon into a unified representation $O$ that serves as conditional guidance for action generation. By integrating features from gesture, audio, and phoneme modalities, this module ensures strong phoneme-level synchronization between speech content and full-body movement.

\textit{GAP Fusion Module:}
The core of the Perception module is our proposed Gesture-Audio-Phoneme (GAP) fusion mechanism, which achieves fine-grained multimodal alignment through hierarchical encoding and cross-modal attention. The module processes each modality through dedicated encoders: gesture components $O_g$ (pose, expression, translation) are encoded via modality-specific transformers, audio features $O_s$ are extracted by a pretrained HuBERT model \cite{hsu2021hubert} and processed through a dual-pathway encoder using Content-Rhythm Attention (CRA) inspired by \cite{liu2024emage}, and phoneme features $O_m$ are captured through a pretrained XPhoneBERT model \cite{nguyen2023xphonebert}. Audio and phoneme features are then aligned through an adaptive gating mechanism, followed by cross-modal attention that integrates the fused speech representation with gesture features. This design enables the model to capture semantic and rhythm cues and establish explicit correspondence between acoustic-linguistic signals and motion dynamics. The resulting multimodal representation $O=\{O_{g},O_{s},O_{m}\}$ provides rich conditional signals that guide the Decision module to generate gesture actions with precise phoneme-level expressiveness and temporal coherence. Detailed architecture specifications are provided in supplementary details.

\subsubsection{Decision.}
The conditional denoising diffusion model is the backbone for learning action policy following \cite{ze20243d}. Starting from a Gaussian noise sequence $a_K$ of length $H$, the network progressively denoises the latent action sequence over $K$ diffusion steps to recover the clean action $a_0$. This reverse diffusion process is mathematically formulated as:
\begin{equation}
a_{k-1} = \alpha_{k}(a_{k} - \gamma_{k}\epsilon_{\theta}(a_{k}, k, O)) + \sigma_{k}\mathcal{N}(0, \mathrm{I}),
\end{equation}
where $\epsilon_{\theta}$ denotes the noise prediction network, while $\alpha_{k}$, $\gamma_{k}$, and $\sigma_{k}$ are schedule-dependent coefficients controlling the noise removal rate. $\mathcal{N}(0, \mathrm{I})$ represents the added stochastic noise for the generative process.

Upon completing $K$ iterations, we obtain the predicted action trajectory $a^{1:N}_0$. To recover the final holistic gesture, these differential actions must be integrated into the gesture state space. The holistic state $g^n_0$ at the $n$-th frame is reconstructed via:
\begin{equation}
g_{0}^{n} = g_{temp} + \sum_{i=1}^{n} \mathcal{G}(a^{i}_0),
\end{equation}
where $g_{temp}$ represents template gesture model. $\mathcal{G}(\cdot)$ is the inverse transformation function, which presented as:
\begin{equation}
    \mathcal{G}(a^{i}) = a^{i} / (\varepsilon_{scaling} \cdot \Lambda_{adaptive})
\end{equation}

\subsubsection{Training Objectives.}
To ensure the synthesized holistic gestures are both probabilistically diverse and physically consistent, we employ a compound objective function that jointly optimizes policy training stability, gesture reconstruction accuracy, and dynamic velocity fidelity.

\begin{table*}
\centering
\setlength{\tabcolsep}{3pt}
\caption{Quantitative comparison of 3DGesPolicy with state-of-the-art methods on BEAT2 dataset. 3DGesPolicy achieves superior performance on all metrics. The units of FGD, DIV, MSE and LVD are $10^{-1}$, 1, $10^{-8}$, and $10^{-5}$ respectively.}
\label{tab:quan}
\begin{tabular}{lcccccc}
\toprule
\multirow{2}{*}{\textbf{Method}} & \multicolumn{2}{c}{\textbf{Modality}} & \multicolumn{4}{c}{\textbf{Metric}} \\
& Expression & Gesture & FGD $\downarrow$ & DIV $\uparrow$ & MSE $\downarrow$ & LVD $\downarrow$ \\
\midrule
FaceFormer \cite{fan2022faceformer} & \ding{51} & \ding{55} & - & - & 7.787 & 7.593 \\
CodeTalker \cite{xing2023codetalker} & \ding{51} & \ding{55} & - & - & 8.026 & 7.766 \\
\midrule
DisCo \cite{liu2022disco} & \ding{55} & \ding{51} & 9.417 & 9.912 & - & - \\
CaMN \cite{liu2022beat} & \ding{55} & \ding{51} & 6.644 & 10.86 & - & - \\
DSG \cite{yang2023diffusestylegesture} & \ding{55} & \ding{51} & 8.811 & 11.49 & - & - \\
\midrule
Habible et al. \cite{habibie2021learning} & \ding{51} & \ding{51} & 9.040 & 8.213 & 8.614 & 8.043 \\
TalkSHOW \cite{yi2023generating} & \ding{51} & \ding{51} & 6.209 & 13.47 & 7.791 & 7.771 \\
DiffSHEG \cite{chen2024diffsheg} & \ding{51} & \ding{51} & 8.986 & 11.91 & 7.665 & 8.673 \\
EMAGE \cite{liu2024emage} & \ding{51} & \ding{51} & 5.512 & 13.06 & 7.680 & 7.556 \\
MambaTalk \cite{xu2024mambatalk} & \ding{51} & \ding{51} & 5.366 & 13.05 & \textbf{6.289} & 6.897 \\
\midrule
\textbf{3DGesPolicy (Ours)} & \ding{51} & \ding{51} & \textbf{3.190} & \textbf{14.03} & 7.020 & \textbf{2.200} \\
\bottomrule
\end{tabular}
\end{table*}

\textit{Diffusion Loss:}
At each training step $k$, A random noise $\epsilon^k \sim \mathcal{N}(0, \mathbf{I})$ is added to the ground truth action sequence $a_0$ to obtain the noisy input $a_k$. The denoising network $\epsilon_\theta$ is trained to predict the added noise conditioned on the observation representation $O$. The diffusion loss is formulated as:
\begin{equation}
\mathcal{L}_{\text{diff}} = \mathrm{MSE}(\epsilon^{k},\epsilon_{\theta}(\overline{\alpha_{k}}a_{0}+\overline{\beta_{k}}\epsilon^{k},k,O)),
\end{equation}
where $\overline{\alpha_{k}}$ and $\overline{\beta_{k}}$ are noise schedules.

\textit{Reconstruction Loss:}
To prevent accumulated drift in long sequences, we explicitly supervise the gesture trajectory by calculating the Euclidean distance between predicted gestures $g^{1:N}$ and ground truth sequence $\hat{g}^{1:N}$:
\begin{equation}
\mathcal{L}_{\text{rec}} = \mathbb{E}_{n}[\frac{1}{N}\sum_{n=1}^{N}{\| g_{0}^{n} - \hat{g}_{0}^{n} \|}^{2}],
\end{equation}
where $g_{0}^n$ and $\hat{g}_{0}^n$ are the predicted and ground truth gestures at frame $n$.

\textit{Velocity Loss:}
To enhance motion smoothness and dynamic fidelity, we apply a velocity loss that captures temporal movement patterns:
\begin{equation}
\mathcal{L} _{\rm{vel}} = \mathbb{E}_{n}[ \frac{1}{N}\sum_{n=1}^{N}{\| (g_{0}^{n-1}-g_{0}^{n})  - (\hat{g}_{0}^{n-1}-\hat{g}_{0}^{n}) \|}^{2} ].
\end{equation}

\textit{Total Objective:}
The final training objective is a weighted sum of the three components:
\begin{equation}
\mathcal{L} = \lambda_1 \mathcal{L}_{\text{diff}} + \lambda_2 \mathcal{L}_{\text{rec}} + \lambda_3 \mathcal{L}_{\text{vel}},
\end{equation}
where $\lambda_1$, $\lambda_2$, and $\lambda_3$ balance the relative importance of each loss term.

\subsubsection{Implementation details.} We implement 3DGesPolicy using the DDIM scheduler with 100 training steps and 10 inference steps. The model is trained for 200 epochs with a batch size of 32 using the AdamW optimizer at an initial learning rate of $1 \times 10^{-5}$. The loss components are balanced with empirically determined weights: $\lambda_1 = 10$, $\lambda_2 = 1$, and $\lambda_3 = 1$. All experiments are conducted on a single NVIDIA RTX 6000 Ada Generation GPU with training completed in 28 hours, 1 minute, and 20 seconds.

\begin{figure*}[t]
  \centering
  \includegraphics[width=0.8\linewidth]{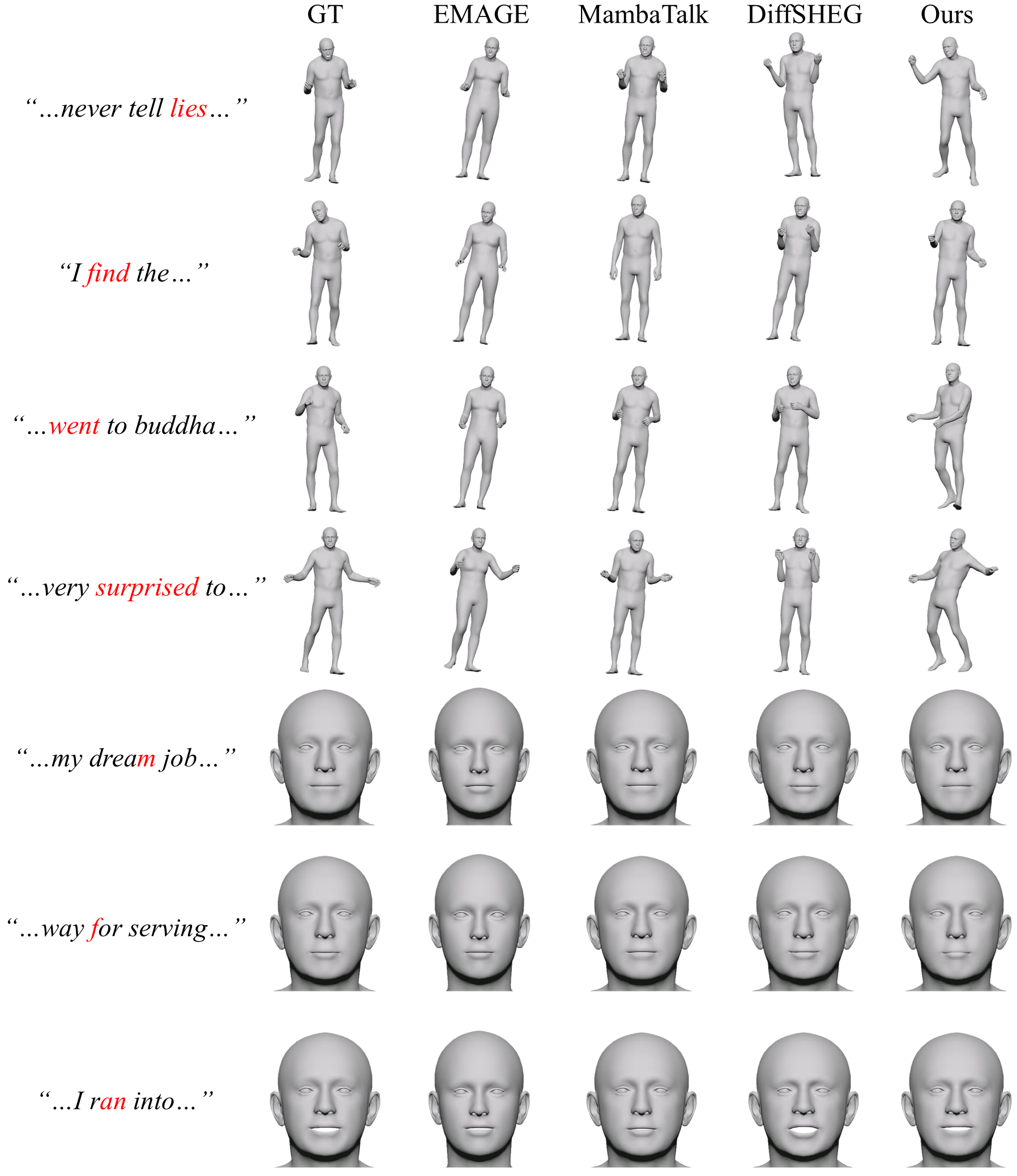}
  \caption{Qualitative comparison of holistic gesture generation on the BEAT2 dataset. Our method produces more expressive and speech-aligned body movements while avoiding static or semantically irrelevant gestures, and achieves superior lip-speech synchronization with accurate capture of subtle phonetic details compared to baseline methods.}
  \label{ges}
\end{figure*}


\section{Experiments}
\label{sec:Experiment}

We present comprehensive experiments to validate the effectiveness of our 3DGesPolicy framework. We conduct both quantitative and qualitative evaluations on the BEAT2 dataset \cite{liu2024emage}, comparing our method against body gesture generation methods \cite{liu2022disco,liu2022beat,yang2023diffusestylegesture}, facial generation methods \cite{fan2022faceformer,xing2023codetalker} and state-of-the-art holistic generation methods \cite{habibie2021learning,yi2023generating,chen2024diffsheg,liu2024emage,xu2024mambatalk} across multiple metrics. To further evaluate our method based on audiovisual user experience, we design a user study that evaluates naturalness, synchronization, and emotional expressiveness through pairwise comparisons. Additionally, we perform ablation studies to analyze the contribution of key components in our framework. Additional results are provided in the \textbf{supplementary material}.

\subsection{Experimental Settings}

\subsubsection{Baseline.}
We compare our 3DGesPolicy with body gesture generation methods \cite{liu2022disco,liu2022beat,yang2023diffusestylegesture} and facial generation methods \cite{fan2022faceformer,xing2023codetalker}, then we mainly compare with the state-of-the-art holistic generation methods \cite{habibie2021learning,yi2023generating,chen2024diffsheg,liu2024emage,xu2024mambatalk}. All methods are evaluated under identical experimental settings to ensure valid comparisons.

\subsubsection{Dataset.}
We train and evaluate our model on the BEAT2 dataset introduced by \cite{liu2024emage}. BEAT2 extends the original BEAT dataset \cite{liu2022beat} with 76 hours of high-quality motion capture data from 30 speakers, standardized into mesh representation with synchronized audio, text, and frame-level semantic annotations \cite{SimLabel,zhang2025unifiedevaluationframeworkmultiannotator}. Following the standard protocol established in \cite{xu2024mambatalk}, we utilize the BEAT2-standard subset and adopt an 85\%/7.5\%/7.5\% split for training, validation, and testing to ensure fair comparison with prior work \cite{liu2024emage,xu2024mambatalk}.

\subsubsection{Evaluation Metric.}
To comprehensively evaluate the quality of generated facial animations, we employ four metrics: Fréchet Gesture Distance (FGD) \cite{yoon2020speech} measures the distributional similarity between generated and ground truth gestures to assess realism. Diversity (DIV) \cite{li2021audio2gestures} measures the gesture variation by computing average L1 distance across different gesture clips. Mean Squared Error (MSE) \cite{xing2023codetalker} quantifies the positional accuracy of facial vertices by computing the average squared distance between predicted and ground truth vertex positions. Lip Vertex Distance (LVD) \cite{yi2023generating} captures the L1 difference between generated and ground truth facial vertices, providing a complementary measure of facial motion quality that is particularly sensitive to lip movement accuracy. The calculation formulas of these metrics are listed in the supplementary material.


\subsection{Quantitative Evaluation}
\label{secsec:quan_eva}
We compare 3DGesPolicy with state-of-the-art methods including expression-only methods \cite{fan2022faceformer,xing2023codetalker}, gesture-only methods \cite{liu2022disco,liu2022beat,yang2023diffusestylegesture}, and holistic generation approaches \cite{habibie2021learning,yi2023generating,chen2024diffsheg,liu2024emage,xu2024mambatalk} in Table~\ref{tab:quan}. Our method achieved superior performance compared with baselines. The significant advantage on FGD and LVD validates 3DGesPolicy's effectiveness in preserving atable and semantically meaningful movements and establishing precise phoneme-level synchronization between speech signals and lip movements through trajectory control.

\begin{table*}[t]
  \centering
  \caption{User study results. Preference percentages for naturalness, synchrony, and emotion. Superior naturalness and high emotion/synchrony scores demonstrate that our method ensures stronger realism while achieving precise semantic synchronization and expressive motion.}
  \label{user}
  \begin{tabular}{lcc|cc|cc}
    \toprule
    \multirow{2}{*}{Methods} & \multicolumn{2}{c}{naturalness (\%) $\uparrow$} & \multicolumn{2}{c}{Synchrony (\%) $\uparrow$} & \multicolumn{2}{c}{Emotion (\%) $\uparrow$} \\
    & Ours & Competitor & Ours & Competitor & Ours & Competitor \\
    \midrule
    Ours \textit{vs.} DiffSHEG \cite{chen2024diffsheg} & \textbf{91.67} & 8.33 & \textbf{58.33} & 41.67 & \textbf{66.67} & 33.33 \\
    Ours \textit{vs.} EMAGE \cite{liu2024emage} & \textbf{83.33} & 16.67 & \textbf{83.33} & 16.67 & \textbf{58.33} & 41.67 \\
    Ours \textit{vs.} MambaTalk \cite{xu2024mambatalk} & \textbf{91.67} & 8.33 & 50.00 & 50.00 & 50.00 & 50.00 \\
    Ours \textit{vs.} GT & 25.00 & \textbf{75.00} & 25.00 & \textbf{75.00} & 0.00 & \textbf{100.00} \\
    \bottomrule
  \end{tabular}
\end{table*}

\begin{table*}[t]
\centering
\caption{Ablation studies on Action, GAP Fusion Module and Phoneme components. The action formulation is crucial for natural motion, while the GAP module and phoneme features are essential for accurate facial animation and lip synchronization.}
\label{tab:abla}
\begin{tabular}{lccc}
\toprule
\textbf{Model Configuration} & FGD $\downarrow$ & MSE $\downarrow$ & LVD $\downarrow$ \\
\midrule
w/o Action & $23.563 \pm 0.481$ & $82.530 \pm 0.269$ & $18.030 \pm 0.405$\\
Sim-GAP Fusion & $3.713 \pm 0.138$ & $12.847 \pm 0.177$ & $2.500 \pm 0.098$ \\
w/o Phoneme & $4.009 \pm 0.125$ & $16.698 \pm 0.242$ & $2.400 \pm 0.076$ \\
\midrule
Full Model & $\textbf{3.190} \pm 0.119$ & $\textbf{7.020} \pm 0.164$ & $\textbf{2.200} \pm 0.088$ \\
\bottomrule
\end{tabular}
\end{table*}

\subsection{Qualitative Evaluation}
We present qualitative comparisons for holistic gesture generation on the BEAT2 template. The upper part of Figure~\ref{ges} shows that our 3DGesPolicy produces natural and diverse gestures that are strongly aligned with speech semantics. By reformulating the task as trajectory control, our method generates motions that are stable and grounded in 3D space, significantly reducing unnatural sliding artifacts compared to existing baselines. Moreover, our policy effectively captures speech content through meaningful gesticulations while avoiding excessive or meaningless movements. For example, our method generates appropriate pointing gestures for words like "went" and "find". It also expresses emotional intensity through dynamic poses for phrases such as "...very surprised to...", whereas baseline methods remain relatively static. The lower part of Figure~\ref{ges} illustrates our superior lip-speech synchronization. Our method accurately reproduces subtle phonetic features, such as complete lip closure for the "m" phoneme and labiodental articulation for "f". Competing methods like EMAGE and MambaTalk often fail to capture these details, resulting in incomplete articulation. These visual results demonstrate that the GAP fusion module establishes precise phoneme-level alignment, which is consistent with our quantitative LVD improvements.


\subsection{User Study}
We conducted a user study through A/B testing to compare our 3DGesPolicy against DiffSHEG \cite{chen2024diffsheg}, EMAGE \cite{liu2024emage}, and MambaTalk \cite{xu2024mambatalk}. Participants were asked to evaluate video pairs based on three dimensions: naturalness, speech synchronization, and emotional expressiveness. From 30 participants, we collected 360 valid assessments. As presented in Table~\ref{user}, our method almost outperforms all competing approaches in every metric, with particularly high scores in naturalness. These results confirm that our action-based trajectory control produces motion patterns that more closely resemble authentic human dynamics in 3D space compared to previous methods. Furthermore, the strong performance in synchronization and emotion validates that our approach achieves superior audio-visual alignment and affective expression.


\subsection{Ablation Study}
We conduct ablation studies on the BEAT2 dataset to validate three core components: action formulation, the GAP fusion module, and phoneme integration. Performance is measured using FGD, MSE, and LVD with units of $10^{-1}$, $10^{-8}$, and $10^{-5}$, respectively. Results are shown in Table~\ref{tab:abla}.

\subsubsection{Action Formulation.}
We compare our action-based trajectory control against a variant using direct gesture regression with absolute poses. The results show a drastic performance drop across all metrics, particularly in FGD and MSE. Validating our core contribution that action-based formulation not only ensures motion continuity through temporal accumulation but also enables the model to learn movement patterns following speech dynamics through trajectory control.

\subsubsection{GAP Fusion Module.}
To evaluate our fusion architecture, we test a simplified version (Sim-GAP) that replaces adaptive gating and cross-modal attention with simple concatenation. We observe notable degradation on facial metrics MSE and LVD, indicating that naive concatenation fails to establish fine-grained alignment. This validates that our GAP module provides the structured representation necessary for precise phoneme-level synchronization.

\subsubsection{Phoneme Integration.}
Finally, we evaluate the contribution of phoneme-level semantic information by removing phoneme features entirely and conditioning only on gesture and audio observations. The results show consistent degradation across all metrics. Validating that phonemes provide critical linguistic timing cues that are essential for both accurate lip-sync and semantic alignment between speech and gestures.


\section{Conclusion}
\label{sec:Conclusion}
We present 3DGesPolicy, a novel trajectory-control framework that reformulates holistic co-speech gesture generation as an action-based diffusion policy problem. By modeling frame-to-frame variations as unified actions and introducing the GAP fusion module for phoneme-level multimodal alignment, our method produces natural gestures with precise lip-speech synchronization. Quantitative and qualitative results on BEAT2 dataset confirm significant improvements in motion quality and expressiveness over state-of-the-art methods. Trajectory control of keypoints on 3D human body gesture may improve our method with fine-grained hand details and finger articulations. Future work will focus on this machenism to enhance hand motion fidelity while maintaining holistic coordination.


\begin{figure*}[t]
  \centering
  \includegraphics[width=0.8\linewidth]{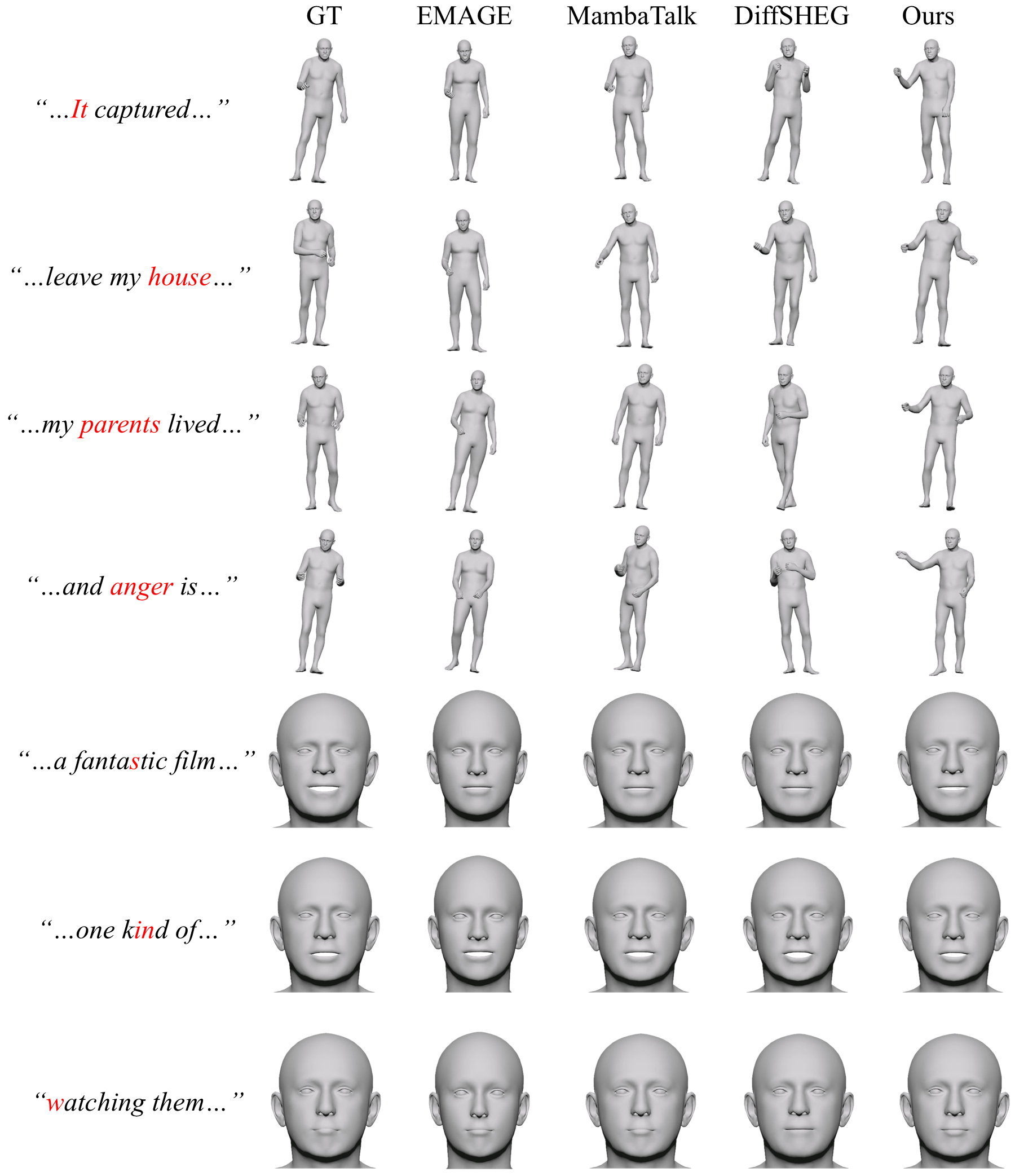}
    \caption{Additional experiment results of gesture generation and facial generation on BEAT2 datasets.}
    \label{ges2}
\end{figure*}

\section{Supplementary Material}
\subsection{Overview}
In this supplementary, we first provide additional experimental results of BEAT2 with the comparison between our approach and other state-of-the-art methods in Sec.~\ref{sec:add}. In Sec.~\ref{sec:eva}, we introduce the calculation formula of metrics. In Sec.~\ref{sec:repro}, we provide the detailed settings of sequence sampler and network architecture of our GAP fusion module for model reproducibility. Finally, we present the details of the user study in Sec.\ref{sec:User_Study}. 

\subsection{Additional Experimental Results}
\label{sec:add}
Due to the limited space in Sec. Experiments, we present additional qualitative comparison results from BEAT2 datasets in Figure~\ref{ges2}. These supplementary results further demonstrate our method's superior performance in natural, expressive and semantically meaningful gesture movements.

The upper part presents additional comparison results for full-body gesture synthesis. Our method consistently generates semantically rich and emotionally expressive body movements that align closely with speech content. In contexts requiring emphasis, such as the words "it" and "anger", our approach produces clear pointing gestures with the hands, while baseline methods fail to capture these meaningful expressions. Similarly, when conveying negative emotions or enumerating items, as in "house" and "parents", other methods generate less expressive movements compared to our results. These observations validate that our trajectory-control framework effectively understands speech semantics and synthesizes motion patterns that naturally reflect the speaker's communicative intent, producing body gestures that are both semantically coherent and emotionally appropriate.

The lower part illustrates additional results for facial expression synthesis, demonstrating our method's superior phoneme-level lip synchronization. During rapid multi-syllable articulations, such as the "s" phoneme in "fantastic", our approach accurately reproduces lip shapes consistent with the ground truth, whereas competing methods produce overly smoothed mouth movements that fail to capture individual syllable articulations. Similarly, for the "w" phoneme in "watching" and the "in" sound in "kind", our method generates lip configurations that closely match the ground truth and exhibit distinctly more accurate articulation compared to baseline approaches. These results confirm the effectiveness of our phoneme-level architectural design in the GAP fusion module, enabling more precise facial motion synthesis with fine-grained audio-visual alignment.

\subsection{Evaluation Metrics}
\label{sec:eva}
Detailed metric calculation description is presented as follows:

\textit{i) Fréchet Gesture Distance (FGD) \cite{yoon2020speech}:}
This metric evaluates the distance of contribution between the ground truth and generated body gestures:
\begin{equation}
    \text{FGD}(\mathbf{g}, \hat{\mathbf{g}}) = \|\mu_r - \mu_g\|^2 + \operatorname{Tr} \left( \Sigma_r + \Sigma_g - 2(\Sigma_r \Sigma_g)^{1/2} \right),
\end{equation}
where $\mu_r$ and $\Sigma_r$ denote the mean and covariance matrix of the latent feature distribution $\mathbf{z}_r$ extracted from real human gestures $\mathbf{g}$, while $\mu_g$ and $\Sigma_g$ represent the corresponding statistics for the latent feature distribution $\mathbf{z}_g$ of the synthesized gestures $\hat{\mathbf{g}}$. Following prior work \cite{liu2024emage,xu2024mambatalk}, we utilize a pretrained autoencoder consisting of a Skeleton CNN (SKCNN)-based encoder and a Full CNN-based decoder to extract these latent features. A lower FGD value indicates that the generated gestures more closely resemble the distribution of authentic human movements.

\textit{ii) L1 Diversity (DIV) \cite{li2021audio2gestures}:} 
This metric quantifies the variability within the generated gesture sequences by computing the average L1 distance across multiple gesture clips. Higher diversity scores reflect greater variance and naturalness in the generated motions. The calculation over $N$ motion clips is formulated as:
\begin{equation}
    \text{DIV} = \frac{1}{2N(N-1)} \sum_{i=1}^{N} \sum_{j=1}^{N} \left\| p^i_t - \hat{p}^j_t \right\|_1 ,
\end{equation}
where $p^i_t$ represents the joint positions at frame $t$ in the $i$-th motion clip. We evaluate diversity across the entire test dataset. Note that when computing joint positions, we set translation to zero, ensuring that L1 Diversity focuses exclusively on local motion patterns.

\textit{iii) Mean Squared Error (MSE) \cite{xing2023codetalker}:} 
This metric measures the positional accuracy of generated facial landmarks by computing the average squared Euclidean distance between predicted and ground truth vertex positions:
\begin{equation}
    \text{MSE} = \frac{1}{n} \sum_{i=1}^{n} (f_i - \hat{f}_i)^2,
\end{equation}
where $n$ denotes the total number of facial vertices, $f_i$ represents the ground truth position of the $i$-th vertex, and $\hat{f}_i$ indicates its predicted position. The summation is performed over all vertices to obtain the average positional error. Lower MSE values indicate higher accuracy in facial motion reconstruction.

\textit{iv) Lip Vertex Distance (LVD) \cite{yi2023generating}:} 
This metric evaluates the synchronization between speech and facial expressions by measuring the velocity difference of lip region vertices. It calculates the average L1 distance between the velocities of generated and ground truth facial vertices:
\begin{equation}
    \text{LVD} = \frac{1}{n} \sum_{i=1}^{n} \left\| f'_i - \hat{f}'_i \right\|_1 ,
\end{equation}
where $n$ denotes the number of vertices in the lip region, $f'_i$ represents the ground truth velocity of the $i$-th vertex, and $\hat{f}'_i$ denotes the velocity of the corresponding vertex in the generated facial expression. The summation is taken over all lip vertices to compute the average absolute velocity difference. Lower LVD values indicate better audio-visual synchronization and more natural lip movements.

\subsection{Model Reproducibility}
\label{sec:repro}
\subsubsection{Sequence Sampler Details.}
To enable efficient policy learning within manageable computational budgets, the sequence sampler partitions long-form sequences into fixed-duration windows called observation horizons $H$. After extensive evaluation experiments, we set $H=16$ to achieve the best performance. Each sampled batch consists of an observation window of length $N_{obs}=8$ for encoding historical context and an action prediction window of length $N_{act}=8$ for generating future gestures. This partitioning ensures the policy learns from locally consistent temporal context while maintaining sufficient history for motion continuity. The sampler processes the gesture sequence $g^{1:N}$, audio sequence $s^{1:N}$, phoneme sequence $m^{1:N}$, and the calculated action sequence $a^{1:N}$ simultaneously to maintain temporal synchronization. This mechanism allows the model to handle variable length inputs by organizing them into standardized batch structures suitable for GPU acceleration.

\subsubsection{GAP Fusion Module Architecture Details.}
The GAP fusion module implements a hierarchical encoding strategy to achieve fine-grained alignment across gesture, audio, and phoneme modalities. This section provides complete architectural specifications for reproducibility.

\textit{i) Modality-Specific Encoders:}
Each modality is processed through dedicated encoders before fusion. For gesture components, we employ three separate Modality Encoder modules for pose ($55 \times 3$ dimensions), expression (100 dimensions), and translation (3 dimensions). Each ModalityEncoder consists of a two-layer MLP projection followed by a Transformer Block for temporal modeling. The audio encoder implements a dual-pathway architecture inspired by \cite{liu2024emage}. After extracting 1024-dimensional features from the pretrained HuBERT model \cite{hsu2021hubert}, we process them through parallel content and rhythm pathways. The content pathway uses a two-layer MLP, while the rhythm pathway employs 1D convolutions with kernel sizes of 5 and 3 to capture temporal patterns. These pathways are then fused using Content-Rhythm Attention (CRA), which computes attention weights to adaptively combine content and rhythm information. For phoneme encoding, we leverage the pretrained XPhoneBERT model \cite{nguyen2023xphonebert} to obtain 768-dimensional contextualized phoneme representations. These features are processed through a Phoneme Encoder consisting of temporal projection layers and 1D convolutions with kernel sizes of 3 and 5 to capture local phoneme patterns, followed by a Transformer Block for sequence modeling. All encoders project their respective inputs to a unified latent dimension of 256.

\textit{ii) Audio-Phoneme Alignment:}
To establish temporal coherence between audio and phoneme modalities, we employ an Audio-Phoneme-Alignment module with an adaptive gating mechanism. This module projects both audio features $O_s$ and phoneme features $O_m$ through separate linear layers, then concatenates them and passes the result through a two-layer MLP with sigmoid activation to compute gate weights $\alpha \in [0,1]$. The aligned speech features are obtained as $O_{speech} = \alpha \cdot O_s + (1-\alpha) \cdot O_m$, where the gate adaptively balances acoustic and linguistic information based on local temporal context.

\textit{iii) Cross-Modal Fusion:}
After encoding, we combine gesture features from pose, expression, and translation through element-wise addition to form unified motion features $O_{motion} = O_{pose} + O_{exp} + O_{trans}$. Both motion features and speech features are augmented with sinusoidal positional encodings to preserve temporal order information. We then apply cross-modal attention where speech features serve as queries and motion features serve as keys and values. This attention mechanism establishes explicit correspondence between acoustic-linguistic signals and motion dynamics. The output is fused with speech features through residual connection: $O_{fused} = \text{CrossAttention}(O_{speech}, O_{motion}) + O_{speech}$. Finally, two stacked TransformerBlocks refine the fused features, and a three-layer MLP projects them to the final observation representation with dimension 1024.

\textit{iv) Implementation Details:}
All Transformer Blocks use 8 attention heads with dropout rate 0.1. The MLP hidden dimension is set to 512 (twice the latent dimension). We use ReLU activation for all intermediate layers except the gating mechanism which uses sigmoid activation. Layer normalization is applied before each attention and feedforward operation following the pre-norm architecture.

\subsection{User Study}
\label{sec:User_Study}
\begin{figure*}[h]
\centering
\includegraphics[width=0.85\linewidth]{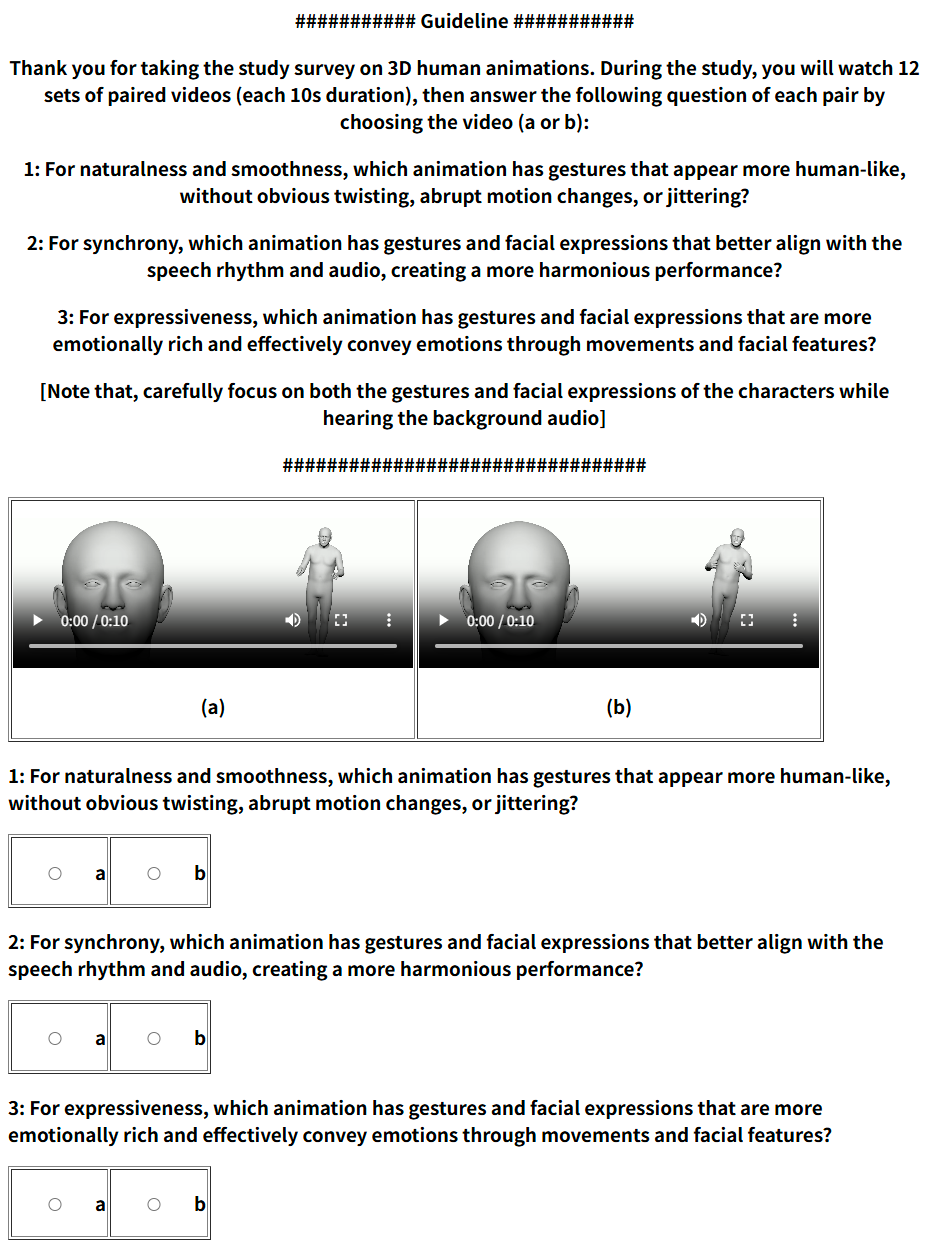}
\centering
\caption{Designed user study interface. Each participant need to answer 12 video pairs and here only one video pair is shown due to the page limit.}
\label{us}
\end{figure*}

The designed user study interface is depicted in Figure~\ref{us}. Participants are asked to make side-by-side comparisons and select the better animation based on their personal preferences. Each video pairs are selected from examples randomly and switch the side randomly between our model and competitors. The completion time for each participant is about 20 minutes, with 12 video pairs and 3 questions for each pair.
The questions that participants are presented with are as follows: (1) For naturalness and smoothness, which animation has gestures that appear more human-like, without obvious twisting, abrupt motion changes, or jittering? (2) For synchrony, which animation has gestures and facial expressions that better align with the speech rhythm and audio, creating a more harmonious performance? (3) For expressiveness, which animation has gestures and facial expressions that are more emotionally rich and effectively convey emotions through movements and facial features?

\clearpage

\end{document}